\documentclass[runningheads]{llncs}
\usepackage{hyperref}
\usepackage[T1]{fontenc}
\usepackage{graphicx,verbatim}
\usepackage{subcaption}
\usepackage{cite} 
\usepackage{amsmath}
\usepackage{amssymb}

\begin{document}
\title{DELST: Dual Entailment Learning for Hyperbolic Image-Gene Pretraining in Spatial Transcriptomics}

%

\author{Xulin Chen \and
Junzhou Huang}
\authorrunning{X. Chen et al.}
%
\institute{University of Texas at Arlington, USA 
 \\
\email{xxc6289@mavs.uta.edu, jzhuang@uta.edu}}

\maketitle              
\begin{abstract}
Spatial transcriptomics (ST) maps gene expression within tissue at individual spots, making it a valuable resource for multimodal representation learning. Additionally, ST inherently contains rich hierarchical information both across and within modalities. For instance, different spots exhibit varying numbers of nonzero gene expressions, corresponding to different levels of cellular activity and semantic hierarchies. However, existing methods rely on contrastive alignment of image-gene pairs, failing to accurately capture the intricate hierarchical relationships in ST data. Here, we propose \textbf{DELST}, the first framework to embed hyperbolic representations while modeling hierarchy for image-gene pretraining at two levels: (1) Cross-modal entailment learning, which establishes an order relationship between genes and images to enhance image representation generalization; (2) Intra-modal entailment learning, which encodes gene expression patterns as hierarchical relationships, guiding hierarchical learning across different samples at a global scale and integrating biological insights into single-modal representations. Extensive experiments on ST benchmarks annotated by pathologists demonstrate the effectiveness of our framework, achieving improved predictive performance compared to existing methods. Our code
and models are available at: \url{https://github.com/XulinChen/DELST}.


\end{abstract}
\section{Introduction}
Histopathology images are essential for disease diagnosis and prognosis, providing insights into tissue morphology and disease progression \cite{Gurcan2009, Bera2019}. However, automated analysis is challenging due to the high resolution of whole-slide images and the complexity of associating cellular morphology with clinical outcomes \cite{7780635, YAO2020101789}. While bulk gene expression profiling \cite{Mortazavi2008, Canzar2016} captures average gene expression across a tissue sample, it lacks spatial resolution. Single-cell RNA sequencing \cite{Zhang2023CellBridge, Stuart2020SingleCell} enables gene expression analysis at the cellular level, yet it disrupts tissue architecture and does not preserve spatial context.

\begin{figure}[htbp]
    \centering
    \begin{subfigure}{0.45\textwidth}
        \centering
        \includegraphics[width=\textwidth]{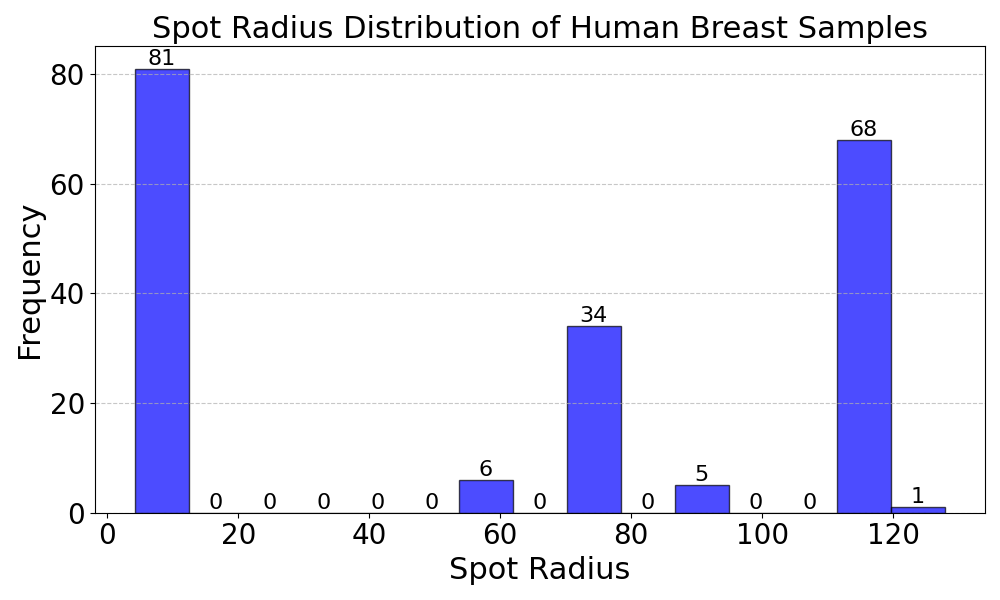}
        \caption{}
        \label{fig:sub1}
    \end{subfigure}
    \begin{subfigure}{0.45\textwidth}
        \centering
        \includegraphics[width=\textwidth]{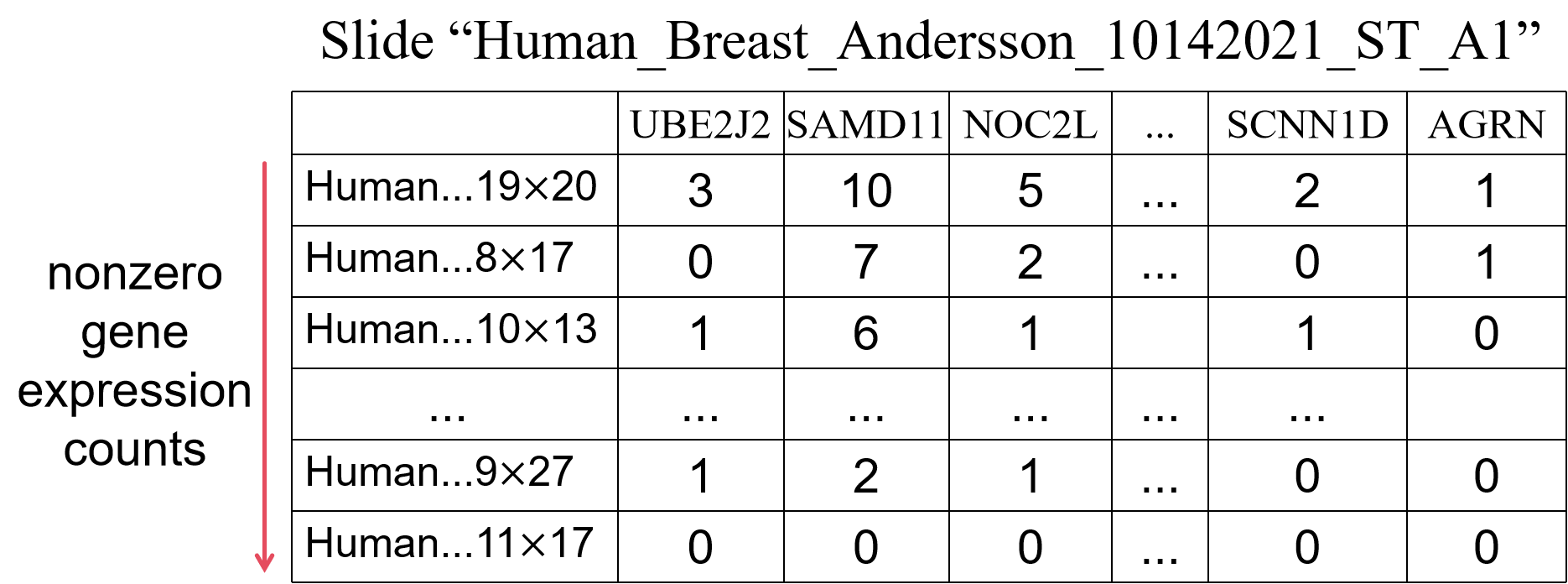}
        \caption{}
        \label{fig:sub3}
    \end{subfigure}
    \caption{\textbf{Characteristics of ST Data}. (a) The spot radius distribution of human breast, the tissue type with the largest sample size in the STimage-1K4M dataset \cite{chen2024stimage1k4m}, spans a wide range. (b) The spots on the same slide are ranked in descending order based on the number of nonzero gene expressions per spot, indicating varying levels of cellular activity across different spots.}
    \label{fig:subfigures}
\end{figure}

Spatial transcriptomics (ST) \cite{Stahl2016Visualization, 10.1126/science.aat5691} is an innovative technology that enables gene expression profiling while preserving spatial context within tissue structures. By providing high-dimensional annotations for each spatial spot within the whole tissue slide, ST facilitates a deeper understanding of tissue organization, cell-cell interactions, and disease progression \cite{Tian2023}. The characteristics of ST data make it a valuable resource for multimodal representation learning in computational pathology. Recent studies have leveraged ST data within contrastive learning frameworks to develop spot-level image-gene pretraining models \cite{chen2024stimage1k4m, jaume2024hest}.

However, leveraging gene-specific information in ST data to pretrain a pathology image encoder remains challenging. Firstly, as an emerging technology, ST data exhibits inconsistencies due to variations in technical platforms \cite{chen2024stimage1k4m}. For example, as shown in Figure \ref{fig:sub1}, the spot radius distribution of human breast in STimage-1K4M dataset \cite{chen2024stimage1k4m}, varies widely. This presents a significant challenge for fine-tuning existing pathology image foundation models. Previous studies extract 112$\times$112 $\mu$m image patches (0.5$\mu$m/px) centered around each spot \cite{jaume2024hest}, which can enhance fine-tuning robustness. However, this approach may also introduce new issues, as images can present more fine-grained details than their corresponding gene expression data. 

Secondly, ST data contains rich biological priors, yet how to integrate them into the image-gene pretraining remains an underexplored problem. For example, as shown in Figure \ref{fig:sub3}, the spots in STimage-1K4M \cite{chen2024stimage1k4m} are ranked in descending order based on the number of nonzero gene expressions per spot. Since gene expression reflects cellular activity within each spot \cite{Rao2021, Yu2024}, this suggests that different spots correspond to varying levels of cellular activity. However, existing methods \cite{chen2024stimage1k4m, jaume2024hest} primarily rely on simple contrastive image-gene alignment and struggle to capture features that represent specific cellular functional activities.

In this paper, we introduce a pioneering hierarchical approach to addressing the aforementioned challenges. Notably, in the field of vision-language contrastive pretraining, prior works \cite{desai2023meru, pal2024} have incorporated explicit hierarchies to optimize image and text encoders. MERU \cite{desai2023meru} maps Euclidean embeddings from image and text encoders onto hyperbolic space, enforcing a 'text entails image' partial ordering through an entailment loss, as text generally conveys broader concepts than images. 

Inspired by previous work \cite{desai2023meru}, we propose DELST, a contrastive learning framework that models hierarchy at both cross-modal and intra-modal levels within hyperbolic space. First, image and gene embeddings are projected from Euclidean to hyperbolic space, which naturally accommodates exponential growth and efficiently represents hierarchical structures \cite{Lee2018}. Building on this, to address spot radius inconsistencies, we introduce cross-modal entailment learning, enforcing the "gene entails image" relationship. Since image patches centered on spots may capture finer details than their corresponding gene expressions, this constraint enhances the generalization of image representations. Additionally, intra-modal entailment learning is introduced by encoding gene expression data into hierarchical relationships. Specifically, since gene expression reflects cellular activity within a spot \cite{Rao2021, Yu2024}, we quantify nonzero gene expression counts (NGEC) for each spot and establish an entailment ordering, where low-NGEC (LNGEC) spots entail high-NGEC (HNGEC) spots. This design enables the model to learn representations that more accurately capture cellular activity across different spots. Our contributions are summarized as follows:
\begin{itemize}
    \item We introduce a novel hierarchical learning approach DELST for image-gene pretraining in ST data, designed to mitigate the impact of imperfections in ST datasets while learning more generalizable representations that effectively capture cellular activities.
    \item We propose Dual Entailment Learning, which enforces both cross-modal (image-gene pair) and intra-modal (ordering relationships across different samples) constraints, optimizing the image and gene encoders through a combination of contrastive loss and entailment loss.
    \item Extensive experiments on ST benchmarks validate our framework, demonstrating improved performance in linear probing.
\end{itemize}

\section{Related Works}

\subsubsection{Multi-Modal Pretraining in Computational Pathology.} 
Recent advances in vision-language pretraining have enabled CLIP-based methods for pathology datasets, such as MI-zero \cite{10203628}, PLIP \cite{huang2023visual}, and CONCH \cite{lu2024avisionlanguage}.
While CLIP-based models excel in representation learning, ST datasets offer finer granularity by incorporating gene expression data. STimage-1K4M \cite{chen2024stimage1k4m} pairs sub-tiles with gene expression profiles for spot-level multi-modal learning, while HEST-1K \cite{jaume2024hest} supports biomarker discovery and gene expression prediction. However, leveraging gene-specific information in ST data to pretrain a pathology image encoder remains an underexplored challenge.

\section{Method}
\begin{figure*}[t]
\centering
\includegraphics[width=\textwidth]{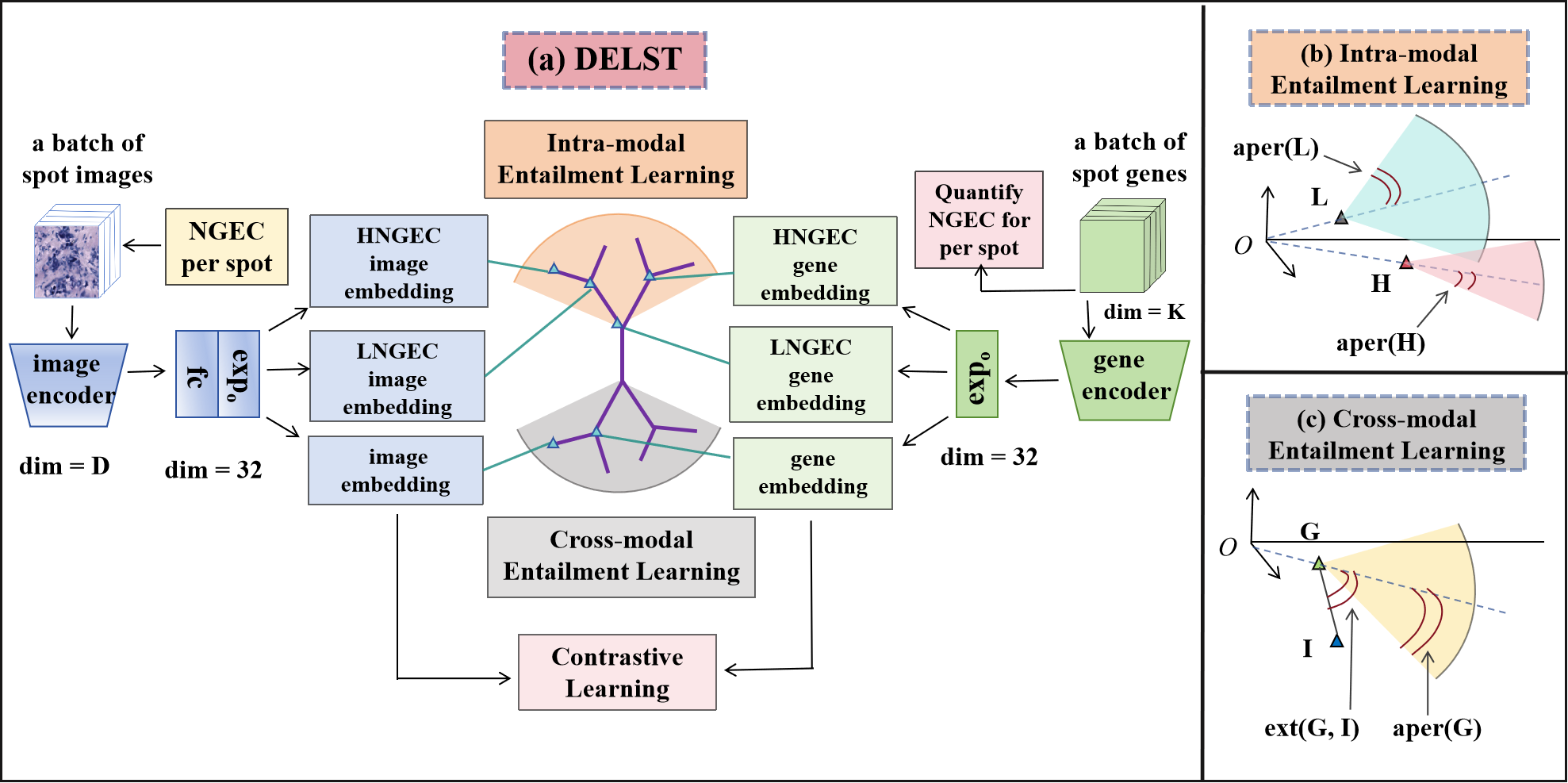}
\caption{\textbf{Overview of DELST. }(a) Spot images and gene expressions are encoded separately and projected into hyperbolic space via the exponential map. DELST enforces cross-modal and intra-modal hierarchies by positioning broader concepts near and finer-grained concepts farther from the hyperboloid’s origin. (b) $\mathbf{H}$ (HNGEC spot) corresponds to a finer-grained hierarchy than $\mathbf{L}$ (LNGEC spot). This intra-modal entailment relationship is independently applied to the gene ($\mathbf{L^{G}}$, $\mathbf{H^{G}}$) and image ($\mathbf{L^{I}}$, $\mathbf{H^{I}}$) modalities. (c) The image embedding $\mathbf{I}$ is pushed to be within the cone projected by its paired gene embedding $\mathbf{G}$.} \label{fig1}
\end{figure*}

We propose DELST, a contrastive and hierarchical learning framework that models cross-modal and intra-modal ordering relationships between image and gene in hyperbolic space (Figure \ref{fig1}a). First, we briefly review hyperbolic geometry concepts, and then introduce our dual entailment learning, designed specifically for ST data characteristics. (Figure \ref{fig1}b and Figure \ref{fig1}c).

\subsubsection{Preliminaries} 
Hyperbolic geometry, characterized by negative curvature and exponential volume growth, differs from Euclidean geometry in its ability to naturally accommodate hierarchical structures, making it well-suited for representing tree-like data\cite{nickel2016learning, krioukov2010hyperbolic}. In DELST, we adopt the Lorentz model to represent hyperbolic space. This model embeds an $n$-dimensional hyperbolic space within the upper sheet of a two-sheeted hyperboloid in $(n+1)$-dimensional spacetime. Every vector $\mathbf{u}\in \mathbb{R}^{n+1}$ is expressed as $[\mathbf{u}_{space}, u_{time}]$, where $\mathbf{u}_{space}\in \mathbb{R}^{n}$ represents the spatial dimensions, and $u_{time}\in \mathbb{R}$ corresponds to the time dimension.

For two vectors $\mathbf{u},\mathbf{v}\in \mathbb{R}^{n+1}$, the Lorentzian inner product is defined as:
\begin{equation}
\begin{split}
\langle \mathbf{u}, \mathbf{v} \rangle_{\mathcal{L}}=\langle \mathbf{u}_{space}, \mathbf{v}_{space} \rangle-u_{time}v_{time}.
\end{split}
\end{equation}
Here, $\langle \cdot, \cdot \rangle$ denotes the Euclidean inner product. The Lorentz model, characterized by a constant curvature $-c$, is defined as the set of vectors:
\begin{equation}
\begin{split}
\mathcal{L}^{n}=\{ \mathbf{u}\in \mathbb{R}^{n+1}:\langle \mathbf{u}, \mathbf{u} \rangle_{\mathcal{L}}=-\frac{1}{c}\},c>0.
\end{split}
\end{equation}
All vectors in this set satisfy the following constraint: 
\begin{equation}
\begin{split}
u_{time}=\sqrt{\frac{1}{c}+\|\mathbf{u}_{space}\|^{2}},
\end{split}
\label{time}
\end{equation}

\subsubsection{Projecting embeddings onto the hyperboloid}
Let the embedding vector from the image or gene encoder, after linear projection, be $\mathbf{p}_{enc}\in \mathbb{R}^{n}$. Following \cite{desai2023meru}, we define the vector $\mathbf{p}=[\mathbf{p}_{enc},0]\in \mathbb{R}^{n+1}$, which lies in the tangent space at the hyperboloid origin $\mathbf{O}$, where $\mathbf{p}_{enc}=\mathbf{p}_{space}$.
To project this vector onto the hyperboloid, we employ the $exponential$ $map$, given by:
\begin{equation}
\begin{split}
\mathbf{u}_{space}=\text{exp}^{c}_{\mathcal{O}}(\mathbf{p}_{space})=\frac{\text{sinh}(\sqrt{c}\|\mathbf{p}_{space}\|)}{\sqrt{c}\|\mathbf{p}_{space}\|}\mathbf{p}_{space}.
\end{split}
\label{exp}
\end{equation}
The time component $u_{time}$ is then computed from $\mathbf{u}_{space}$ using Eqn. \ref{time}, ensuring that the resulting vector $\mathbf{u}$ always lies on the hyperboloid.

\subsubsection{Image Processing and Encoder}
To accommodate variations in spot radius, we extract $224 \times 224$-pixel image patches centered around each spot. These patches are used to fine-tune the image encoder of pretrained CONCH \cite{lu2024avisionlanguage}, which is the ViT-B/16 visual-language foundation model. A fully connected layer then maps the output into a 32-dimensional latent space. Subsequently, the exponential map in Eqn. \ref{exp} is used to obtain the projected embedding $\mathbf{I}$.

\subsubsection{Gene Processing and Encoder}
To handle the high dimensionality of gene expression data, we employ three gene selection strategies, resulting in a final input of $K$ genes: (1) Highly variable genes (HVG) selected independently for each slide ($K=128$); (2) HVGs from overlapping genes across slides (overlap-HVG) ($K=100$); (3) HVGs from overlapping genes across slides, excluding those with zero counts in more than 90$\%$ of spots (e-overlap-HVG) ($K=100$). The first two strategies follow \cite{chen2024stimage1k4m}, while the third follows \cite{jaume2024hest}. After gene selection, the spot gene expression data is passed through a fully connected layer, transforming the $K$-dimensional input into a 32-dimensional embedding, following \cite{chen2024stimage1k4m}. Finally, the exponential map is applied to obtain the projected embedding $\mathbf{G}$.

\subsubsection{Contrastive Learning}
For image-gene pretraining, we employ contrastive learning to align image and gene features. Given a batch of $N$ spots, each associated with an image feature $\mathbf{I}$ and a gene feature $\mathbf{G}$, the contrastive loss is defined as:
\begin{equation}
\begin{split}
L_{cont}(\mathbf{I},\mathbf{G})=-\frac{1}{N}\sum_{i=1}^{N}\text{log}\frac{\text{exp}(\text{sim}(\mathbf{I}_{i},\mathbf{G}_{i})/\tau)}{\sum_{j=1}^{N}\text{exp}(\text{sim}(\mathbf{I}_{i},\mathbf{G}_{j})/\tau)},
\end{split}
\label{cl}
\end{equation}
where $\text{sim}(\cdot,\cdot)$ denotes the cosine similarity, and $\tau$ is a temperature parameter. 

\subsubsection{Cross-modal Entailment Learning (CMEL)}
Due to variations in spot radii (Figure \ref{fig:sub1}), extracting fixed-size image patches can introduce inconsistencies between image and gene expression, as images capture more contextual information while gene expression is hierarchically more general. To address this, we enforce a gene-entails-image ordering in the representation space using hyperbolic entailment cones. Let $\mathbf{G}$ and $\mathbf{I}$ denote the embeddings of a gene-image pair. The entailment cone for $\mathbf{G}$ is defined by its half-aperture \cite{desai2023meru}:
\begin{equation}
\begin{split}
\text{aper}(\mathbf{G})=\text{sin}^{-1}(\frac{2K}{\sqrt{c}\|\mathbf{G}_{space}\|}),
\end{split}
\label{aper}
\end{equation}
where $K=0.1$ controls boundary conditions. The aperture shrinks as $\|\mathbf{G}_{space}\|$ increases, positioning general concepts closer to the origin and specific ones farther away. To enforce entailment, $\mathbf{I}$ is pushed to be within $\mathbf{G}$'s cone. The exterior angle (Figure \ref{fig1}c) is measured based on the formulation by Desai et al. \cite{desai2023meru}:
\begin{equation}
\begin{split}
\text{ext}(\mathbf{G}, \mathbf{I})=\text{cos}^{-1}(\frac{I_{time}+G_{time}c\langle \mathbf{G},\mathbf{I}\rangle_{\mathcal{L}}}{\|\mathbf{G}_{space}\|\sqrt{(c\langle \mathbf{G},\mathbf{I}\rangle_{\mathcal{L}})^{2}-1}}).
\end{split}
\label{ext}
\end{equation}
If $\text{ext}(\mathbf{G}, \mathbf{I})$ exceeds $\text{aper}(\mathbf{G})$, $\mathbf{I}$ is adjusted using the following loss function:
\begin{equation}
\begin{split}
L_{ent\_cross}(\mathbf{G},\mathbf{I})=\frac{1}{N}\sum_{i=1}^{N}\text{max}(0, \text{ext}(\mathbf{G}_i,\mathbf{I}_i)-\text{aper}(\mathbf{G}_i)).
\end{split}
\label{cross_entail}
\end{equation}
This enforces the cross-modal ordering structure between gene and image in the latent space.

\subsubsection{Intra-modal Entailment Learning (IMEL)}
For IMEL, we aim to integrate biological priors from ST data into the representation space of images and genes, extending hierarchical learning from local image-gene pairs to a global sample perspective. Concretely, considering that gene expression serves as an indicator of cellular activity within a spot \cite{Rao2021, Yu2024}, we quantify each spot’s nonzero gene expression counts (NGEC) to indicate its activity and semantic hierarchy. 

Given a batch of $N$ spots, we select the top-$Q$ spots with the highest NGEC (HNGEC) and the top-$Q$ with the lowest NGEC (LNGEC), enforcing the pairwise entailment "LNGEC entails HNGEC". Here, LNGEC corresponds to a more general level closer to the origin, while HNGEC represents a more specific level farther from the origin (Figure \ref{fig1}b). This is enforced through the following loss function:
\begin{equation}
\begin{split}
L_{ent\_intra}(\mathbf{L},\mathbf{H})=\frac{1}{Q^2}\sum_{i=1}^{Q}\sum_{j=1}^{Q}\text{max}(0, \text{aper}(\mathbf{H}_j)-\text{aper}(\mathbf{L}_i)).
\end{split}
\label{intra_entail}
\end{equation}
The intra-modal entailment relationship is applied independently to both gene and image modalities. The total IMEL loss $L_{ent\_intra}$ is computed as the average of the two losses $L_{ent\_intra}(\mathbf{L}^{\mathbf{G}},\mathbf{H}^{\mathbf{G}})$ and $L_{ent\_intra}(\mathbf{L}^{\mathbf{I}},\mathbf{H}^{\mathbf{I}})$. This formulation induces intra-modal hierarchical ordering across different spots in the latent space, enabling the model to learn biologically meaningful representations.

We compute the final loss as a weighted sum of the contrastive loss and entailment losses:
\begin{equation}
\begin{split}
L_{final}=L_{cont}+\lambda L_{ent\_cross}+\beta L_{ent\_intra}.
\end{split}
\label{final_loss}
\end{equation}

\section{Experiments and Results}
\subsubsection{Training Dataset}
Due to different genes measured across datasets and batch effects,  STimage-1K4M \cite{chen2024stimage1k4m} limited its image-gene pretraining study to human brain samples from Maynard et al. \cite{Maynard2020TranscriptomescaleSG}. To demonstrate the effectiveness of our DELST on large-scale data, we conduct experiments using two tissue types: (1) human breast, the most abundant tissue type in STimage-1K4M dataset \cite{chen2024stimage1k4m}, with 195 WSIs and 209,201 spots; (2) human brain samples from \cite{Maynard2020TranscriptomescaleSG}, the same tissue type used in STimage-1K4M \cite{chen2024stimage1k4m}, with 12 WSIs and 47,681 spots.

\subsubsection{Evaluation Benchmark and Metric}
We benchmark the performance of image-gene pretraining models on pathologist-annotated datasets from STimage-1K4M \cite{chen2024stimage1k4m} using linear probing for image classification. The benchmark consists of 24 human breast slides with 29,569 spot-level classifications and 12 human brain slides with 47,329 spot-level classifications. For linear probing, we follow the procedure in STimage-1K4M \cite{chen2024stimage1k4m}. We first extract image embeddings using different models. A simple linear classifier is then trained on 80$\%$ of the annotated spots (train:validation:test $=$ 8:1:1), using five different seeds and image embeddings from different models. For evaluation, we use the mean F1 score, following \cite{chen2024stimage1k4m}. The image encoder trained on human breast tissue is evaluated on the human breast benchmark, while the image encoder trained on human brain data (Maynard et al. \cite{Maynard2020TranscriptomescaleSG}) is evaluated on the human brain benchmark.

\subsubsection{Implementation}
We finetune the last 3 layers of the image encoder in CONCH \cite{lu2024avisionlanguage}. For a fair comparison, all baselines and DELST use 224$\times$224-pixel patches centered on the spot as input. Since STimage-1K4M \cite{chen2024stimage1k4m} crops spot regions as input, we reproduce better baseline results than those reported in \cite{chen2024stimage1k4m}. We use the Adam optimizer with a learning rate of $5 \times 10^{-5}$ and weight decay of 0.2. The batch size is 1024 and the temperature $\tau$ in contrastive learning is 0.07. All models are trained for 15 epochs in a single H100 GPU. For hyperparameters, both $\lambda$ and $\beta$ are set to 0.1 (Eqn. \ref{final_loss}), and $Q$ in IMEL is 150. We tested $[0.1,0.2]$ for $\lambda$, $\beta$, and $[50,100,150,200]$ for $Q$, selecting values based on the results of the validation set from the benchmark.

\subsubsection{Baselines}
We categorize the baselines into two groups: (1) non-finetuned models, including CLIP \cite{radford2021learning}, PLIP \cite{huang2023visual} and CONCH \cite{lu2024avisionlanguage}, which serve as frozen encoders to extract image embeddings for individual ST spots, (2) finetuned models, specifically CONCH-ft, where the last three layers of the image encoder are finetuned. The finetuning approach follows STimage-1K4M \cite{chen2024stimage1k4m}, employing multimodal contrastive learning.
For both non-finetuned models and fine-tuned models, we evaluate the 512-dimensional image embedding.

\begin{table}[t]
\centering
\caption{F1 scores from linear probing using different image encoders.}
\label{tab1}
\begin{tabular}{|c|c|c|c|}
\hline
Model &  Gene Selection Strategy & Human Breast & Human Brain\\
\hline
CLIP \cite{radford2021learning} &  \textbackslash{} & 0.682$\pm$0.100 & 0.567$\pm$0.045\\
PLIP \cite{huang2023visual} &  \textbackslash{} & 0.719$\pm$0.095 & 0.620$\pm$0.040\\
CONCH \cite{lu2024avisionlanguage} & \textbackslash{} & 0.746$\pm$0.082 & 0.641$\pm$0.032\\
\hline
CONCH-ft \cite{chen2024stimage1k4m} & HVG & 0.751$\pm$0.086 & 0.663$\pm$0.035\\
\textbf{DELST (Ours)} & HVG & \textbf{0.772$\pm$0.083} & \textbf{0.697$\pm$0.031}\\
\hline
CONCH-ft \cite{chen2024stimage1k4m} & overlap-HVG & 0.766$\pm$0.082 & 0.668$\pm$0.036\\
\textbf{DELST (Ours)} & overlap-HVG & \textbf{0.775$\pm$0.083} & \textbf{0.678$\pm$0.027}\\
\hline
CONCH-ft \cite{chen2024stimage1k4m} & e-overlap-HVG & 0.755$\pm$0.085 & 0.668$\pm$0.034\\
\textbf{DELST (Ours)} & e-overlap-HVG & \textbf{0.784$\pm$0.083} & \textbf{0.674$\pm$0.026}\\
\hline
\end{tabular}
\end{table}

\begin{table}[t]
\centering
\caption{Ablation study evaluating the F1 score of DELST. Each result is averaged over the three gene selection strategies.}
\label{tab2}
\begin{tabular}{|c|c|c|c|c|}
\hline
Variants &  CMEL & IMEL & Human Breast & Human Brain\\
\hline
CONCH-ft &  &  & 0.758$\pm$0.084  & 0.666$\pm$0.035\\
DELST w/o IMEL & $\checkmark$ &  & 0.765$\pm$0.093 & 0.676$\pm$0.030\\
\textbf{DELST}  & $\checkmark$ & $\checkmark$ & \textbf{0.777$\pm$0.083} & \textbf{0.683$\pm$0.028}\\
\hline
\end{tabular}
\end{table}

\subsubsection{Comparison}
As shown in Table \ref{tab1}, finetuned models consistently outperform non-finetuned models. After finetuning, CONCH-ft further improves performance, demonstrating the effectiveness of image-gene contrastive learning. Compared to CONCH-ft, DELST improves performance across all gene selection strategies, highlighting the effectiveness of entailment learning, which enforces both cross-modal and intra-modal hierarchical constraints, leading to more biologically meaningful representations and enhancing expressivity of embeddings.

\subsubsection{Ablation Study}
Table \ref{tab2} presents the ablation study results. Introducing CMEL improves performance over CONCH-ft, showing its effectiveness in enhancing the generalization of image features. Further incorporating IMEL leads to consistent improvements, confirming its role in inducing hierarchical structure in embeddings and learning biologically meaningful features.

\section{Conclusion}
We propose DELST, a hierarchical learning framework enforcing cross-modal and intra-modal entailment constraints in ST data. Experiments on ST benchmarks show that DELST consistently outperforms baselines, demonstrating the effectiveness of Dual Entailment Learning.

%

%
%
\bibliographystyle{splncs04}

%

\end{document}